\documentclass[acmsmall,screen,nonacm]{acmart}
\usepackage[utf8]{inputenc}

\acmJournal{PACMPL}
\acmVolume{1}
\acmNumber{CONF} % CONF = POPL or ICFP or OOPSLA
\acmArticle{1}
\acmYear{2018}
\acmMonth{1}
\acmDOI{} % \acmDOI{10.1145/nnnnnnn.nnnnnnn}
\startPage{1}

\setcopyright{none}
\usepackage{booktabs}   %% For formal tables:
\usepackage{subcaption} %% For complex figures with subfigures/subcaptions
\usepackage{alltt}
\usepackage{graphicx}
\usepackage{listings}
\usepackage{mathpartir}
\usepackage{marginalia}

\newcommand{\mit}[1]{\mathit{#1}}
\newcommand{\mtt}[1]{\texttt{#1}}

\newcommand{\jsobj}[1]{\ensuremath{\texttt{\{}{#1}\texttt{\}}}}
\newcommand{\jsarr}[1]{\ensuremath{\texttt{[}{#1}\texttt{]}}}

\newcommand{\jinja}[1]{\ensuremath{\mtt{\textdollar}\texttt{\{}{#1}\texttt{\}}}}

\newcommand{\pdldef}[2]{#1 \texttt{:} {#2} }

\newcommand{\pdlmodel}[3]{\ensuremath{{\mtt{model:}{#1}\mtt{,}\mtt{input:}{#2}\mtt{,}\mtt{parameters:}{#3}}}}
\newcommand{\pdlcode}[2]{\ensuremath{{\mtt{code:}{#1}\mtt{,}\mtt{lang:}{#2}}}}

\newcommand{\pdldata}[1]{\ensuremath{{\mtt{data:}{#1}}}}
\newcommand{\pdltext}[1]{\ensuremath{{\mtt{text:}{#1}}}}
\newcommand{\pdlarray}[1]{\ensuremath{{\mtt{array:}{#1}}}}
\newcommand{\pdlobject}[1]{\ensuremath{{\mtt{object:}{#1}}}}
\newcommand{\pdllastOf}[1]{\ensuremath{{\mtt{lastOf:}{#1}}}}
\newcommand{\pdlif}[3]{\ensuremath{{\mtt{if:}{#1}\mtt{,} \mtt{then:}{#2}\mtt{,} \mtt{else:}{#3}}}}
\newcommand{\pdlrepeat}[3]{\ensuremath{{\mtt{repeat:}{#1}\mtt{,} \mtt{num\_iterations:}{#2}\mtt{,} \mtt{join:}{#3}}}}
\newcommand{\pdlrepeatuntil}[3]{\ensuremath{{\mtt{repeat:}{#1}\mtt{,} \mtt{until:}{#2}\mtt{,} \mtt{join:}{#3}}}}
\newcommand{\pdlfor}[3]{\ensuremath{{\mtt{for:}{#1}\mtt{,} \mtt{repeat:}{#2}\mtt{,} \mtt{join:}{#3}}}}
\newcommand{\pdlread}[1]{\ensuremath{{\mtt{read:}{#1}}}}
\newcommand{\pdlinclude}[1]{\ensuremath{{\mtt{include:}{#1}}}}
\newcommand{\pdlfunction}[2]{\ensuremath{{\mtt{function:}{#1}\mtt{,}  \mtt{return:}{#2}}}}

\newcommand{\pdlcall}[2]{\ensuremath{{\mtt{call:}{#1}\mtt{,} \mtt{args:}{#2}}}}

\definecolor{keyword}{HTML}{37AC4A} 
\definecolor{jinja2}{HTML}{0070C0}
\definecolor{comment}{HTML}{7F7F7F}
\definecolor{generated}{HTML}{37AC4A} 
\definecolor{toolout}{HTML}{A51DFF}

\lstdefinelanguage{pdl}{
  xleftmargin=4mm,
  numbers=left,
  numbersep=2mm,
  numberstyle=\tiny\color{gray},
  basicstyle=\fontsize{8}{9}\ttfamily,
  basewidth=0.5em,
  alsoletter={:},
  morekeywords={array:,as:call:,code:,content:,contribute:,data:,def:,defs:,description:,else:,for:,function:,if:,include:,input:,join:,lang:,lastOf:,message:,model:,multiline:,num_iterations:,object:,parameters:,parser:,pdl_context:,read:,repeat:,return:,role:,spec:,text:,then:,until:,with:},
  keywordstyle=\color{keyword}\bf,
  morestring=[s]{$\{}{\}},
  stringstyle=\color{jinja2},
  emphstyle=\color{jinja2},
  showstringspaces=false,
  morecomment=[l]{\#},
  commentstyle=\color{comment}\it,
}

\begin{document}
\begin{abstract}
Large language models~(LLMs) have taken the world by storm by making
many previously difficult uses of AI feasible.
LLMs are controlled via highly expressive textual prompts and return
textual answers.
Unfortunately, this unstructured text as input and output makes
LLM-based applications brittle.
This motivates the rise of prompting frameworks, which mediate between
LLMs and the external world.
However, existing prompting frameworks either have a high learning
curve or take away control over the exact prompts from the developer.
To overcome this dilemma, this paper introduces the Prompt Declaration
Language~(PDL).
PDL is a simple declarative \emph{data-oriented} language that puts
prompts at the forefront, based on YAML.
PDL works well with many LLM platforms and LLMs.
It supports writing interactive applications that call LLMs and tools,
and makes it easy to implement common use-cases such as chatbots, RAG,
or agents.
We hope PDL will make prompt programming simpler, less brittle, and
more enjoyable.

\end{abstract}

\title{PDL: A Declarative Prompt Programming Language}
\author{Mandana Vaziri}
\authornote{IBM Research}
\author{Louis Mandel}
\authornotemark[1]
\author{Claudio Spiess}
\authornote{UC Davis}
\author{Martin Hirzel}
\authornotemark[1]
\authorsaddresses{}
\maketitle

\section{Introduction}
Large language models~(LLMs) have made great advances, demonstrating
the ability to perform a wide range of useful tasks.
As LLMs are controlled via natural-language \emph{prompts}, prompt
engineering has emerged as an ad-hoc approach to improve
accuracy~\cite{white_et_al_2023}.
Even more capabilities can be unlocked with prompting patterns such as
in-context learning~\cite{brown_et_al_2020}, chaining multiple LLM
calls~\cite{chase_et_al_2022}, retrieval-augmented
generation~(RAG)~\cite{lewis_et_al_2020}, tool use~\cite{schick_et_al_2023},
program-aided language models~(PAL)~\cite{gao_et_al_2023}, and
agents~\cite{yao_et_al_2023}.
Unfortunately, while powerful, LLMs remain brittle: they sometimes
hallucinate, or even fail to comply with expected syntax and types.

Prompting frameworks~\cite{liu_et_al_2023} make it easier for
developers to use LLMs and associated prompting patterns while
ameliorating their brittleness.
Some, such as LangChain~\cite{chase_et_al_2022} and
AutoGen~\cite{wu_et_al_2023}, do so via bespoke features for popular
patterns such as RAG or agents.
Unfortunately, this bespokeness takes control over basic prompts away
from users and forces them to learn many complex framework features.
In contrast, low-level prompting frameworks, such as
Guidance~\cite{microsoft_2023} and
LMQL~\cite{beurerkellner_fischer_vechev_2023}, provide more control
with syntax and types.
Unfortunately, they require users to program in
imperative languages such as Python or TypeScript.
At the other end of the spectrum, frameworks such as
DSPy~\cite{khattab_et_al_2023} and Vieira~\cite{li_et_al_2024} avoid
hand-written prompts altogether by automatically generating them.
Unfortunately, this takes away even more control from the developer.
The problem thus becomes how to make LLM programming less brittle
while keeping it simple and keeping the developer in the driver's seat.

To tackle this problem, we turned to tried-and-true programming language
design ideas.
The principle of \emph{orthogonality} advocates for a small set of simple
features that compose in powerful ways~\cite{vanwijngaarden_et_al_1977}.
Being orthogonal, or at right angles, here means being irredundant and
avoiding exceptional cases as far as possible.
For prompting frameworks, orthogonality is a way to avoid bespoke features.
Next, developers need to struggle less with brittleness if the language
checks \emph{types} and \emph{roles}~\cite{huggingface_2023} to enforce
structure by construction.
One remaining tension is harder to tackle: on the one hand, we want to
give developers control over the exact prompts, but on the other hand,
we want a simple \emph{declarative} language.
To this end, we settled on a \emph{data-oriented language}, which puts
prompts at the forefront by intentionally blurring the line between
programs~(e.g.\ for chaining and tools) and data~(for prompts).
This is inspired by the old idea of code as data~\cite{mccarthy_1960},
as well as by seminal work on programming without
tiers~\cite{cooper_et_al_2006}.

This paper introduces the Prompt Declaration Language~(PDL), an
orthogonal and typed data-oriented language.
Unlike other prompting languages that are embedded in imperative
languages, PDL is based on YAML~\cite{benkiti_et_al_2004}.
YAML is a data serialization format that is both human-readable~(by
promoting a nice and simple syntax for unstructured strings) while
also providing structure~(by being JSON-compatible).
Variables %\fTBD{Louis: why "Variables"?}
in PDL also hold JSON values and can optionally be typed with
JSON Schema~\cite{pezoa_et_al_2016}.
PDL is currently implemented by an interpreter, and the interpreter
performs dynamic type checking. %\fTBD{Louis: dynamic type checking is kind of orthogonal}
One benefit of representing programs as data is that it
facilitates program transformations~\cite{mernik_heering_sloane_2005},
such as for optimization.
Rendering programs in a data representation format even facilitates
PDL programs that generate PDL programs with LLMs, similar to
PAL~\cite{gao_et_al_2023}.

PDL programs comprise blocks~(YAML objects), where each block appends
data to the prompt context.
This mental model is a natural fit for prompting techniques such as
chatbots or agents: program execution implicitly builds up a
conversation or trajectory, without necessitating explicit plumbing.
This context then becomes the input to the next LLM call.
PDL supports local LLMs, and LLMs hosted by various providers, including but not
limited to open-source Granite models~\cite{abdelaziz_et_al_2024,granite_3_0}
on IBM watsonx\footnote{\url{https://www.ibm.com/watsonx}}
and on Replicate\footnote{\url{https://replicate.com/}}.
PDL provides control constructs for looping and conditionals,
as well as functions and file includes for modularity.
PDL adopts Jinja2~\cite{ronacher_2008} expressions to template not
just prompts but entire programs.

This paper gives a quick overview of PDL by means of an introductory
example~(Section~\ref{sec:overview}), followed by a tour of the
language~(Section~\ref{sec:lang}).
It describes the tooling for running and editing PDL
programs~(Section~\ref{sec:tooling}) and offers case studies
illustrating more uses of PDL~(Section~\ref{sec:eval}).
Finally, the paper discusses related work~(Section~\ref{sec:related})
and concludes~(Section~\ref{sec:conclusion}).
PDL is open-source and available at
\url{https://github.com/IBM/prompt-declaration-language}.
Overall, PDL is a simple yet powerful new language for LLM
prompt programming.

\section{Overview}\label{sec:overview}
This section gives an overview of PDL features by means of
a chatbot example.
A PDL program executes a sequence of \emph{blocks}, each of which
generates data that it contributes to the background context.
There are different kinds of blocks, capable of generating data in
different ways: model calls, reading data from stdin or a file,
directly creating various kinds of JSON data, and executing code.
In addition, there are a variety of control blocks (if-then-else, for,
and repeat) that let PDL users express rich data pipelines and AI
applications.

\begin{figure}
\begin{minipage}[b]{0.52\textwidth}
\begin{lstlisting}[emph={question}]
- read:
  contribute: [context]
  message: |
    What is your query?
- repeat:
    text:
      - model: watsonx/ibm/granite-13b-chat-v2
        parameters:
          stop: ["\n\n"]
      - def: question
        read:
        contribute: [context]
        message: |

          Enter a query or say "quit" to exit.
  until: ${question == "quit"}
\end{lstlisting}
\centerline{(a) Code}
\end{minipage}
\hfill
\begin{minipage}[b]{0.47\textwidth}
{\small\input{code/chatbot_output.tex}}\\[2mm]
\centerline{(b) Interpreter trace}
\end{minipage}
\caption{\label{chatbot}Simple Chatbot in PDL}
\end{figure}

Fig.~\ref{chatbot}(a) shows the PDL code for a simple chatbot.
The \lstinline{read:} block on \mbox{Lines 1--4} prints a message
asking the user to enter a query, which it reads from stdin.
Fig.~\ref{chatbot}(b) shows an execution trace of the same program.
For instance, the user might ask `What's a language salad?'.
To avoid duplication, the `\lstinline{contribute: [context]}' clause
puts the user response into the background context but not the result
(printed on stdout).

The \lstinline{repeat:}\hspace*{-1mm}\lstinline{until:} block on \mbox{Lines 5--16} has one
nested \lstinline{text:} block, which in turn has a sequence of two
nested blocks.
The \lstinline{text:} block turns the results of its nested blocks
into strings and concantenates them.
The \lstinline{model:} block on \mbox{Lines 7--9} calls an LLM,
using the accumulated context so far as the prompt.
In the first loop iteration, that context comprises only two lines
`What is your query?' and `What's a language salad?'.
The `\lstinline{stop: [\n\n]}' model parameter causes the LLM to stop
producing tokens after generating two consecutive newline characters.
The LLM interpreter prints LLM outputs in green; Fig.~\ref{chatbot}(b)
shows that in this example, the LLM produced
`\textcolor{generated}{A language salad is [...]}'.
The \lstinline{read:} block on \mbox{Lines 10--15} prints a message
using YAML's multi-line string syntax starting with a vertical
bar~(\lstinline{|}).
This example illustrates how PDL keeps prompts at the forefront while
making them readable and giving the developer precise control.
The interpreter trace on the right shows the user entering `Say it as
a poem!', which Line~10 on the left uses to define variable
\lstinline[emph={question}]{question} and Line~12 appends to the
context.
The \lstinline{until:} clause on Line~16 specifies a Jinja2
expression `\lstinline!${question == "quit"}!' as the loop termination
condition.
PDL embeds Jinja2 templates using `\lstinline!${...}!' syntax rather
than `\lstinline!{{...}}!' because the latter interacts poorly with
YAML, where curly braces are special characters.

In the second loop iteration, the context includes the effect of the
first loop iteration.
Hence, the second execution of the \lstinline{model:} block sees
the output from its first execution and can paraphrase it as a poem,
`\textcolor{generated}{In a world where many tongues [...]}' in
Fig.~\ref{chatbot}(b).
Finally, in this example, during the second execution of the
\lstinline{read:} block the user entered `quit', causing the loop to
terminate.
Now that we have seen a few common PDL blocks in action
(\lstinline{read:}, \lstinline{repeat:}, \lstinline{text:}, and
\lstinline{model:}), we can proceed to Section~\ref{sec:lang}, which
describes the remaining blocks and language features.

\section{Language}\label{sec:lang}
\begin{figure*}
  \centerline{\includegraphics[width=\textwidth]{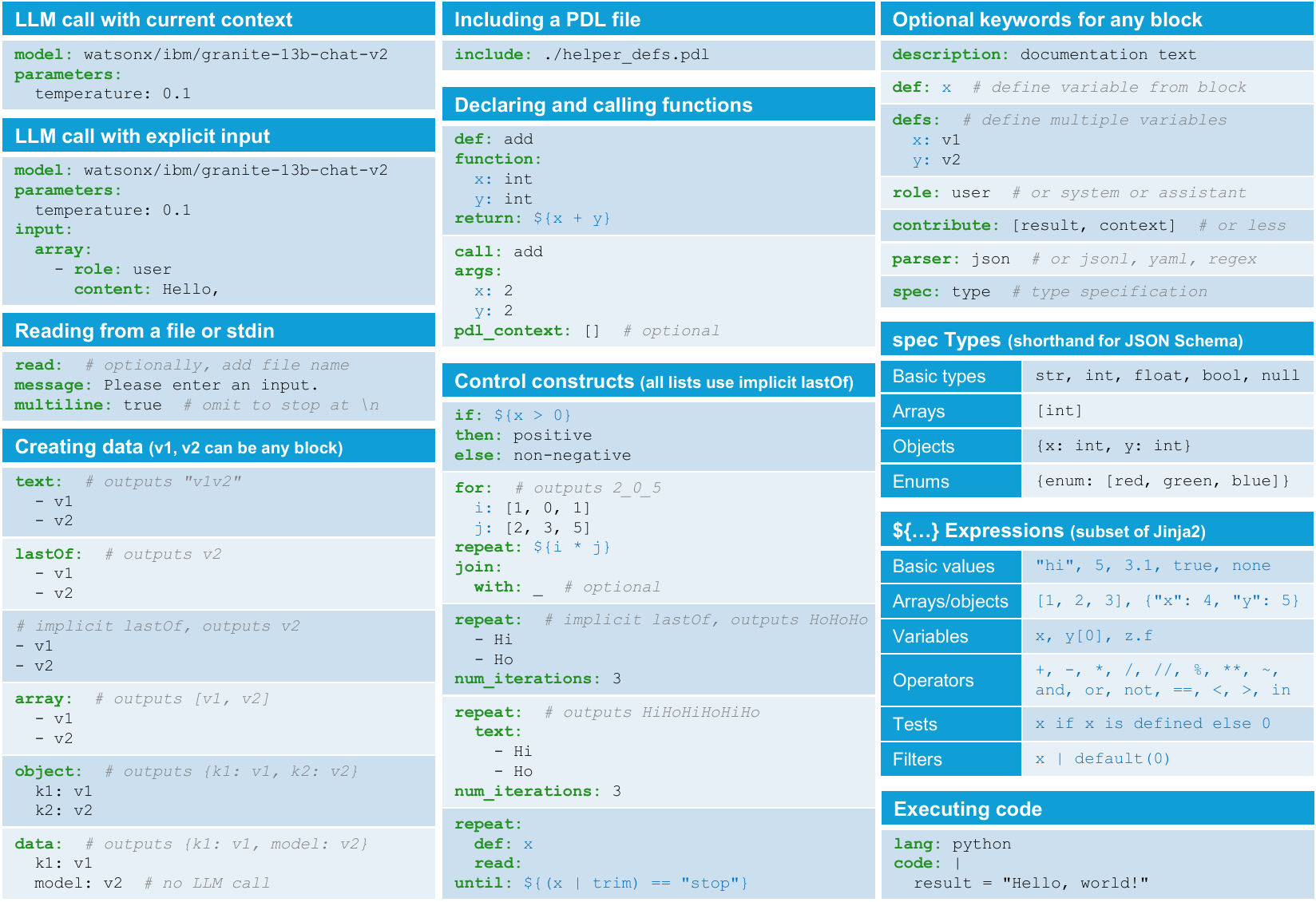}}
  \caption{\label{fig:quick_ref}PDL Quick Reference}
\end{figure*}

PDL is a language embedded into YAML such that every PDL program is a
valid YAML document following the PDL schema\footnote{\url{https://ibm.github.io/prompt-declaration-language/dist/pdl-schema.json}}.
Fig.~\ref{fig:quick_ref} is a quick reference of PDL, and this
section explains it using grammar rules.
A program is a block or a list of blocks where blocks are expressions
or structured blocks, as expressed by the following grammar rules:
\begin{quote}\small$\begin{array}{lcl}
    \mit{pdl} & ::= &\
        \mit{block}
        \mid
        \jsarr{\mit{block}\mtt{,} \dots\mtt{,} \mit{block}}
    \\
    \mit{block} & ::= &\
    \mit{expression} \mid \mit{structured\_block}
\end{array}$\end{quote}

\noindent
All grammar rules in this section use YAML's flow-style syntax
(e.g., \mbox{$\jsarr{\mit{block}\mtt{,} \dots\mtt{,} \mit{block}}$}).
The same PDL code can also be rendered in YAML's block-style syntax,
e.g.:
\begin{quote}\begin{lstlisting}[mathescape,xleftmargin=0mm,numbers=none,aboveskip=0pt,belowskip=0pt]
- $\mit{block}$
$\dots$
- $\mit{block}$
\end{lstlisting}\end{quote}

Each block has a block body, with keywords indicating the block kind
(e.g., \mtt{model} or \mtt{read}).
There are 15~kinds of block bodies (optional fields are annotated with
a question mark):
\begin{quote}\small$\begin{array}{lcl}
    \mit{block\_body} & ::= &\phantom{\mid}
    \pdlmodel{\mit{expression}}{^?\mit{pdl}}{^?\mit{expression}}
    \\ &&\mid
    \pdlread{\mit{file}}\mtt{,}\mtt{message:}^? \mit{string}\mtt{,}\mtt{message:}^? \mit{bool}
    % \\ &&\mid
    % \pdlapi{\mit{url}}{\mit{api}}{\mit{pdl}}
    % \\ &&\mid
    % \pdlget{\mit{x}}
    % \\ &&\mid
    % \pdldatap{\mit{json}}{\mit{string}}
    \\ &&\mid
    \pdltext{\mit{pdl}}
    \\ &&\mid
    \pdllastOf{\mit{pdl}}
    \\ &&\mid
    \pdlarray{\mit{pdl}}
    \\ &&\mid
    \pdlobject{\mit{pdl}}
    \\ &&\mid
    \pdldata{\mit{json}}
    \\ &&\mid
    \pdlinclude{\mit{file}}
    \\ &&\mid
    \pdlfunction{\mit{args}}{\mit{pdl}}
    \\ &&\mid
    \pdlcall{f}{\mit{args}}
    \\ &&\mid
    \pdlif{\mit{expression}}{\mit{pdl}}{^?\mit{pdl}}
    \\ &&\mid
    \pdlfor{\mit{args}}{pdl}{^?\mit{join}}
    \\ &&\mid
    \pdlrepeat{\mit{pdl}}{\mit{n}}{^?\mit{join}}
    \\ &&\mid
    \pdlrepeatuntil{\mit{pdl}}{\mit{expression}}{^?\mit{join}}
    \\ &&\mid
    \pdlcode{\mit{pdl}}{\mit{string}}
\end{array}$\end{quote}

We already saw \lstinline{model:} and \lstinline{read:} blocks
in the previous section.
A \lstinline{model:} block calls an LLM.
The prompt comes from the current context, unless the optional
\lstinline{input:} field is specified, in which case it comes from there.
The optional \lstinline{parameters:} configure the model inferencing
behavior.
A \lstinline{read:} block reads input from a file or from stdin if no
file name is specified.
The optional \lstinline{message:} is displayed to the user and the
optional \lstinline{multiline:} field determines whether to stop at
newline.

There are five kinds of blocks for creating data: \lstinline{text:},
\lstinline{lastOf:}, \lstinline{array:}, \lstinline{object:}, and
\lstinline{data:}.
Fig.~\ref{fig:quick_ref} illustrates them on simple examples.
A plain list of blocks without a keyword behaves like
\lstinline{lastOf:}.
The difference between an \lstinline{object:} block and a
\lstinline{data:} block is that the PDL interpreter ignores PDL
keywords in the latter, treating them like plain JSON fields instead.

For modularity, PDL supports \lstinline{include:} blocks and functions.
An \lstinline{include:} block opens the PDL program at the given
relative path and adds its output at the place where it occured.
Function arguments have the following syntax:
\begin{quote}\small$\begin{array}{lcl}
    \mit{args} & ::= &\
    \jsobj{\pdldef{x}{\mit{expression}}\mtt{,} \dots\mtt{,}\pdldef{x}{\mit{expression}}}
\end{array}$\end{quote}

\noindent
Each $\pdldef{x}{\mit{expression}}$ maps an argument name to either
a type specification (in a \lstinline{function:} definition) or
a value (in a function \lstinline{call:}).
The \lstinline{return:} keyword provides the function body, which can
have nested blocks; Fig.~\ref{fig:quick_ref} shows a simple case where
it is just a Jinja2 expression.
The optional \lstinline{pdl_context:} keyword can reset the background
context for the duration of a call, e.g.\ to the empty context
\lstinline{[]}.

There are three kinds of blocks for control constructs:
\lstinline{if:}, \lstinline{for:}, and different flavors of
\lstinline{repeat:}.
They can have nested blocks or simple expressions, and if they contain
a list of blocks, that list implicitly behaves like
\lstinline{lastOf:}.
If the \lstinline{lastOf:} behavior is not intended, a common idiom is
to wrap the body of a loop in a \lstinline{text:} block,
or to combine loop iteration results with the \lstinline{join:} keyword:
\begin{quote}\small$\begin{array}{lcl}
    \mit{join} & ::= &\
    \mtt{as:}^? (\mtt{text} \mid \mtt{array} \mid \mtt{lastOf})\mtt{,} \mtt{with:}^? \mit{string}
\end{array}$\end{quote}

Each of the 15~kinds of block bodies described above can be orthogonally
composed with zero or more optional keywords that work for any block:
\begin{quote}\small$\begin{array}{lcl}
    \mit{structured\_block} & ::= &\
      \begin{array}[t]{@{}l@{~}l@{~}r@{}}
      \texttt{\{}
        & \mit{block\_body}\mtt{,}
      \\& \mtt{description:}^? \mit{string}\mtt{,}
      \\& \mtt{def:}^? \mit{x}\mtt{,}
      \\& \mtt{defs:}^? \mit{defs}\mtt{,}
      \\& \mtt{role:}^? \mit{string}\mtt{,}
      \\& \mtt{contribute:}^? \mit{contribute}\mtt{,}
      \\& \mtt{parser:}^? \mit{parser}\mtt{,}
      \\& \mtt{spec:}^? \mit{type}
    % \\& \mtt{fallback:}^? \mit{pdl}\mtt{,}
      & \texttt{\}}
      \end{array}
\end{array}$\end{quote}

A \lstinline{description:} is a special comment.
A \lstinline{def:} assigns the result of the block to a variable; we
already saw an example for that in Line~10 of Fig.~\ref{chatbot}.
In contrast, \lstinline{defs:} creates multiple variable definitions,
each with its own name $x$ and a value given by a nested PDL program:
\begin{quote}\small$\begin{array}{lcl}
    \mit{defs} & ::= &\
    \jsobj{\pdldef{x}{\mit{pdl}}\mtt{,} \dots\mtt{,}\pdldef{x}{\mit{pdl}}}
\end{array}$\end{quote}

A \lstinline{role:} causes the data resulting from a block to be
decorated with a role, such as `user', `assistant', or `system'.
When PDL calls a chat model, it follows common practice of modern chat
APIs and passes not a flat string as the prompt, but rather, a
sequence of {\small\texttt{\{content:str, role:str\}}} pairs.
Then, the model API applies a model-specific chat template, which
flattens that sequence by inserting the appropriate control tokens for
that model~\cite{huggingface_2023}.
This gives PDL programs some degree of model-independence.
If a block does not have an explicit \lstinline{role:}, it defaults to
`assistant' for model blocks and to `user' for all other blocks.
Inner nested blocks have the same role as their outer enclosing block.
In future work, we also plan to leverage roles for privilege-based
security.

The \lstinline{contribute:} keyword can specify a (possibly empty)
subset of the two destinations `result' or `context'.
By default, every block contributes to both its own result and the
background context for later LLM calls.
Line~2 of Fig.~\ref{chatbot} showed an example of limiting the
contribution of a block to just the context to declutter the output.

The \lstinline{parser:} keyword makes it possible for a block that
would ordinarily just produce a flat string~(e.g., an LLM call) to
instead produce structured data.
The supported parsers are json, yaml, regex, and jsonl.
The \lstinline{spec:} keyword can specify a type.
PDL's types are a subset of JSON Schema~\cite{pezoa_et_al_2016},
with shorthand syntax for simple commonly used types illustrated
in Fig.~\ref{fig:quick_ref}.
For example, type `\lstinline!{questions: [str], answers: [str]}!'
is an object with two fields \lstinline{questions} and
\lstinline{answers}, both of which hold arrays of strings.
Section~\ref{sec:eval} will illustrate how \lstinline{parser:}
and \lstinline{spec:} can work together.
Future work will also leverage these keywords for
constrained decoding~\cite{scholak_schucher_bahdanau_2021}.

An atomic block is an expression:
\begin{quote}\small$\begin{array}{lcl}
    \mit{expression} & ::= &\
    \mit{bool}
    \mid
    \mit{number}
    \mid
    \mit{string}
    \mid
    \jinja{jinja\_expression}
    \mid
    \mit{string\_expression}
\end{array}$\end{quote}

\noindent
Expressions can be basic values, Jinja2 expressions~\cite{ronacher_2008},
or strings containing Jinja expressions.
Jinja2 is a convenient way for specifying prompt templates, where parts
of a prompt are hardcoded and others are filled in from expressions.
But PDL takes the use of Jinja2 further, by letting developers
template not just individual prompts, but entire chains of model calls
and other blocks.
While we refer the reader to the Jinja2 documentation for an
exhaustive list of possible expressions, Fig.~\ref{fig:quick_ref}
briefly lists the most common ones.
PDL adopts only Jinja2 expressions, not Jinja2 statements such as
{\small\verb|{% if .. %}|} or {\small\verb|{% for .. %}|}, because
those are redundant with PDL's own \lstinline{if:} and \lstinline{for:}.

Last but not least, PDL has a \lstinline{code:} block that allows it
to execute code in a given programming language (at the time of
writing, only Python is supported).
The next section will describe PDL tooling, including the interpreter,
which provides a sandboxing feature to reduce risks associated with
executing arbitrary code.
To learn more, see the tutorial linked from PDL's github repository.

%% \section{Semantics}
%% \input{semantics.tex}

\section{Tooling}\label{sec:tooling}
PDL comes with tools for making PDL programs easy to write, run, and
understand.

First and foremost, the PDL \emph{interpreter} is an execution engine
with a command-line interface, as one would expect from a scripting
language.
The interpreter supports a streaming mode, where LLM outputs become
visible incrementally as they are being produced, for a more
interactive chat experience.
The interpreter also supports sandboxing, which causes it to launch in
a container, recommended when executing LLM-generated actions or code.

The PDL \emph{IDE support} enhances VSCode, making it easier to write
PDL code via syntax highlighting, auto-complete, tooltips for PDL
keywords, and error checking.
These capabilities are, in part, driven by the PDL meta-schema i.e.,
the JSON schema that defines what constitutes valid PDL.

The {\small\verb|%%pdl|} \emph{cell magic} enhances Jupyter Notebooks
so developers can write code cells directly in PDL.
That way, hosted notebook platforms can serve as a simple
playground for interactively exploring prompts.
Given multiple PDL code cells in the same notebook, later cells can
use variables defined in earlier cells.
Furthermore, the background context for later cells is continued from
earlier cells; when not desired, developers can override
this behavior via {\small\verb|%%pdl --reset-context|}.

The PDL \emph{live document visualizer} shows the concrete execution
trace of a PDL program with colored nested boxes, similar to
typical figures in papers or blog posts about LLM prompting.
Then, the user can select one of the boxes to display the
corresponding PDL code, similar to how spreadsheet cells show data,
but the user can select them to inspect the formula that created that
data.
This live view is a way to let users quickly understand concrete data,
and then move from that to understanding the code that produced it.

Finally, PDL has an \emph{SDK} (software development kit), which is a
small Python library for calling from Python into PDL.
This is useful for extending larger Python applications to use
prompt-based programs, such as agents.
As discussed in Section~\ref{sec:lang}, a PDL file can contain Python
in \lstinline{code:} blocks.
When developing larger applications with PDL, we have found it useful
to keep these to a few lines of code, by defining a function in a
separate Python file and then calling it from PDL.
A good practice is to pass data from PDL to Python and vice versa as
JSON objects.
Optionally, this can be type-checked using the \lstinline{spec:}
keyword in PDL, and TypedDict or Pydantic on the Python side,
as illustrated in the next section in Fig.~\ref{rag}.

\section{Case Studies}\label{sec:eval}
We already saw a simple PDL chatbot example in Section~\ref{sec:overview}.
This section illustrates slightly more advanced use-cases for PDL:
RAG, agents, and generating PDL from PDL.

\subsection{Retrieval-Augmented Generation}\label{sec:rag}
\begin{figure}
\begin{minipage}[b]{0.53\textwidth}
\begin{lstlisting}[language=pdl,emph={test_query:,retrieved:,few_shot_sample:}]
text:
- lang: python
  code: |
    import rag_mbpp
    PDL_SESSION.mbpp = rag_mbpp.initialize()
    result = ""
- defs:
    test_query: >-
      Write a python function to remove first and
      last occurrence of a given character from
      the string.
    retrieved:
      lang: python
      spec: [{query: str, answer: str}]
      code: |
        import rag_mbpp
        result = rag_mbpp.retrieve(
            PDL_SESSION.mbpp, "${test_query}", 5
        )
  text: >
    Given the text after "Q:", generate a Python
    function after "A:".

    Here are some examples, complete the last one:
- for:
    few_shot_sample: ${retrieved}
  repeat: |
    Q: ${few_shot_sample.query}
    A: ```${few_shot_sample.answer}```
- |-
    Q: ${test_query}
    A: 
- model: watsonx/ibm/granite-3-8b-instruct
  parameters:
    stop: ["Q:", "A:"]
\end{lstlisting}
\centerline{(a) PDL code}
\end{minipage}
\hfill
\begin{minipage}[b]{0.46\textwidth}
\begin{lstlisting}[language=python]
from typing import TypedDict
import datasets
from sklearn.feature_extraction.text \
    import TfidfVectorizer

def initialize():
    train_in = datasets.load_dataset(
        "mbpp", "sanitized", split="train"
    )
    corpus = [row["prompt"] for row in train_in]
    tfidf = TfidfVectorizer().fit(corpus)
    def embed(text):
        sparse_result = tfidf.transform(
            raw_documents=[text]
        )
        return sparse_result.toarray().flatten()
    train_em = train_in.map(
        lambda row: {"em": embed(row["prompt"])}
    )
    vec_db = train_em.add_faiss_index("em")
    return vec_db, embed

QA = TypedDict("QA", {"query":str,"answer":str})
def retrieve(mbpp, query, n: int) -> list[QA]:
    vec_db, embed = mbpp
    key = embed(query)
    nearest = vec_db.get_nearest_examples(
        "em", key, n
    )
    queries = nearest.examples["prompt"]
    answers = nearest.examples["code"]
    return [
        {"query": q, "answer": a}
        for q, a in zip(queries, answers)
    ]
\end{lstlisting}
\centerline{(b) Python code}
\end{minipage}
\caption{\label{rag}RAG example in PDL}
\end{figure}

Retrieval-augmented generation, or \emph{RAG}, works by first
retrieving relevant context, then adding that to the prompt for a
model to generate an answer~\cite{lewis_et_al_2020}.
Fig.~\ref{rag}(a) shows a PDL program that uses RAG to retrieve
few-shot samples for a code-generation task.
The \lstinline{code:} block in \mbox{Lines 2--6} uses Python to
initialize a vector database for the training split of the MBPP
dataset of ``mostly basic Python programs''~\cite{austin_et_al_2021}.
It uses a Python function defined in Fig.~\ref{rag}(b), together with a
\lstinline{PDL_SESSION} special variable that enables it to carry
state to a later code block.
\mbox{Lines 8--11} of Fig.~\ref{rag}(a) initialize variable
\lstinline[emph={test_query}]{test_query}
with a natural-language request for Python code to be generated.
\lstinline{Lines 12-19} initialize variable
\lstinline[emph={retrieved}]{retrieved} with the five most similar
samples from the training data.

\mbox{Lines 20--24} add an instruction to the context,
\mbox{Lines 25--29} add the few-shot samples to the context,
and \mbox{Lines 30--32} add the test query to the context.
The \lstinline{for:} loop on Line~25 is an idiomatic way to use
PDL for generating data, in this case, for in-context learning.
Finally, \mbox{Lines 33--35} call a Granite~3 model~\cite{granite_3_0}
with the accumulated context, causing it to generate the Python
function requested by the test query.
While this is a simple example, we have also used PDL with
Codellm-Devkit~\cite{krishna_et_al_2024}, which performs static
analysis on source code from various programming languages to retrieve
other relevant context when prompting LLMs for coding tasks.

\subsection{ReAct Agent}\label{sec:react}
\begin{figure}
\begin{minipage}[b]{0.55\textwidth}
\begin{lstlisting}[language=pdl,emph={thought,action,observation}]
text:
- read: react_few_shot_samples.txt
- |

  When was the discoverer of the Hudson River born?
- repeat:
    text:
    - def: thought
      model: watsonx/ibm/granite-34b-code-instruct
      parameters:
        stop: ["Act:"]
        include_stop_sequence: true
    - def: action
      model: watsonx/ibm/granite-34b-code-instruct
      parameters:
        stop: ["\n"]
      parser: json
      spec: {name: str, arguments: {topic: str}}
    - def: observation
      if: ${ action.name == "Search" }
      then:
        text:
        - "Obs: "
        - lang: python
          code: |
            import wikipedia
            query = "${ action.arguments.topic }"
            result = wikipedia.summary(query)
  until: ${ action.name != "Search" }
\end{lstlisting}
\centerline{(a) Code}
\end{minipage}
\hfill
\begin{minipage}[b]{0.44\textwidth}
{\small\input{code/react_output.tex}}\\[2mm]
\centerline{(b) Interpreter trace}
\end{minipage}
\caption{\label{react}ReAct agent in PDL}
\end{figure}

An LLM-based \emph{agent} lets an LLM select and
configure \emph{actions}, executes those actions in an
\emph{environment}, and feeds the outputs from the actions back to the
LLM as \emph{observations}.
There are different patterns for such agents, such as
ReAct~\cite{yao_et_al_2023} and ReWOO~\cite{xu_et_al_2023}.
An action is an LLM-based \emph{tool call}~\cite{schick_et_al_2023},
and an agent chains together multiple tool calls in a dynamic
LLM-directed sequence.
The ambition is to make AI-based applications less
prescriptive and more goal-driven.
Moreover, when something goes wrong with an action, agents can use
the observation as feedback to recover.

Fig.~\ref{react} shows a PDL example of a simple ReAct agent.
The core of ReAct is a think-act-observe loop, which manifests in
the code as variable definitions for
\lstinline[emph=thought]{thought}~(Line~8),
\lstinline[emph=action]{action}~(Line~13), and
\lstinline[emph=observation]{observation}~(Line~19).
The \lstinline[emph=thought]{thought} is model-generated natural language
e.g.,\ `\textcolor{generated}{I need to search the discoverer of the
  Hudson River, find when he was born}'
in the interpreter trace in Fig.~\ref{react}(b).
The \lstinline[emph=action]{action} is model-generated JSON to match the
tool-use training data of Granite models~\cite{abdelaziz_et_al_2024}.
Lines 17 and~18 on the left ensure that the LLM output is parsed as
JSON and conforms to a \lstinline!{name,arguments}! schema, and the
interpreter trace on the right shows that the model indeed generates
such an object.
That makes it possible to access fields of the object with Jinja2,
such as \lstinline!${ action.arguments.topic }! on Line~27.
The \lstinline[emph=thought]{observation} is generated by the
environment, in this case, Python code calling Wikipedia.
As discussed in Section~\ref{sec:tooling}, for cases like this that
involve running code (partially) generated by an LLM, we recommended
PDL's sandboxing feature.

The interpreter trace in Fig.~\ref{react}(b) shows that this execution
had two iterations of the agentic loop.
While this is a simple example, we have also used PDL to implement
a code editing agent that we used as part of a submission\footnote{\url{https://github.com/swe-bench/experiments/tree/main/evaluation/lite/20241016_IBM-SWE-1.0}}
to the SWE-bench Lite leaderboard~\cite{jimenez_et_al_2024}.
This submission was the first to resolve 23.7\% of instances with only
open-source models, which is higher than any previous results with
open-source models, and in the same ball-park as frontier models.

\subsection{Generating PDL from PDL with LLMs}\label{sec:pal}
%\begin{alltt}\textcolor{red}{TODO}\scriptsize
% - for GSM
%- also mention LISP~\cite{mccarthy_1960},
%  PAL
%\end{alltt}

\begin{figure}
\begin{minipage}[t]{0.55\textwidth}
\vspace*{-4mm}
\begin{lstlisting}[language=pdl,emph={demos:,raw:,PDL,tennis_balls,bought_balls,RESULT}]
defs:
  demos:
    data:
      text: 
        |
        ...
	
        Question: Roger has 5 tennis balls.
	He buys 2 more cans of tennis balls. 
        Each can has 3 tennis balls.
	How many tennis balls does he have now?

        Answer: 
        ```
        text:
        - "Roger started with\n"
        - def: tennis_balls
          data: 5
        - "\ntennis ball.\n"
        - "2 cans of 3 tennis balls each is\n"
        - lang: python
          def: bought_balls
          code: result = 2 * 3
        - "\ntennis balls.\n" 
        - "The result is: \n"
        - lang: python
          def: RESULT
          code: result = ${ tennis_balls } + ${ bought_balls }
        ```
    raw: true
text:
  - model: watsonx/meta-llama/llama-3-70b-instruct
    def: PDL
    input: 
      text:
        - ${ demos.text }
        - "Question: ${ question }"
  - lang: python
    code: | 
      from pdl.pdl import exec_str
      s = """${ PDL }"""
      pdl = s.split("```")[1]
      result = exec_str(pdl)
    def: RESULT
\end{lstlisting}
\caption{\label{gsm}PDL meta generation}
\end{minipage}
\hfill
\begin{minipage}[t]{0.44\textwidth}
\input{code/gsmhard.tex}
\caption{\label{gsmhard}GSMHard sample problematic data point}
\end{minipage}
\end{figure}

The previous sections showed examples of how a human developer can
use PDL to encode different prompting patterns.
This section turns to LLMs and shows how they can also be used to generate
PDL. This meta PDL generation is helpful when LLMs need to create
a plan for solving a problem, for example as part of an agentic
workflow. Traditionally, such plans are just text or JSON or
Python code. With PDL, these plans can be a composition of
model and code calls that are fully executable.
This section explores using PDL meta generation for the GSMHard
dataset\footnote{\url{https://huggingface.co/datasets/reasoning-machines/gsm-hard}}.

GSMHard is a harder version of GSM8k, which consists of grade-school
math problems that require simple arithmetic or symbolic
reasoning. GSMHard contains an input which is a math problem statement,
together with an output which is Python code that solves the
problem. We implemented the PAL~\cite{gao_et_al_2023} approach but instead of
generating Python code, we ask an LLM to generate PDL. The
textual chain-of-thought is represented as PDL text blocks, and
arithmetic is done using PDL code blocks.

Fig.~\ref{gsm} shows a PDL program that generates PDL code and executes
it all in the same program. The \lstinline[emph=demos]{demos} variable holds few-shot
samples designed to teach a model how to generate PDL code. On Line~32,
a model call block uses these samples, together with a
\lstinline[emph=question]{question}, which is a free variable, as input. The result
is a PDL program to solve the question. Line~38 extracts the PDL program
and executes it in Python. This program is applied to the
GSMHard dataset, where \lstinline[emph=question]{question} is filled with input
questions.

This experiment resulted in the discovery that 10\% of the GSMHard
dataset is actually incorrect in the sense that the ground truth is
inconsistent with the question that was asked. Fig.~\ref{gsmhard}
shows an example of such an inconsistency. Using PDL
helped in this discovery because the generated PDL code is
human-readable, so we were able to easily check data points that did not
match the ground truth and found that the ground truth is incorrect
in some cases. We used an LLM to cover the entire dataset and
systematically pick examples that seemed inconsistent. We then
manually filtered the result to remove false positives and
identified 10\% of data points that present this problem.

\section{Related Work}\label{sec:related}
A recent survey defines a \emph{prompting framework} as a layer that
manages, simplifies and facilitates the interaction between LLMs and
users, tools, or other models~\cite{liu_et_al_2023}.
The survey highlights that a major pitfall of prompting frameworks is
increasingly steep learning curves.

Perhaps the most popular prompting framework today is
LangChain~\cite{chase_et_al_2022}, whose extensive features make it
both powerful and complex.
Avoiding this complexity is the main motivation for
MiniChain~\cite{rush_2023}, which offers fewer, simpler features that
can be composed for advanced use-cases.
However, both LangChain and MiniChain are Python frameworks, making
them less declarative, since developers must write imperative code.
PDL has a similar motivation to MiniChain, but goes a step further, by
adopting YAML instead of Python as its foundation.

Like other prompting frameworks, PDL aims to make LLMs less brittle.
Guidance~\cite{microsoft_2023} is a Python-based framework that
provides more structure but is more low-level than LangChain.
Similarly, LMQL~\cite{beurerkellner_fischer_vechev_2023} is a
domain-specific language embedded in Python that leverages types and
constrained decoding.
PDL takes some inspiration from the latter in how it intersperses
prompts and programs, but unlike LMQL, relies less on imperative
Python code.
Crouse et al.\ use finite state machines to formally specify the
internals of various agentic loops~\cite{crouse_et_al_2024}; while
fascinating, this work does not introduce a full-fledged prompting
language.

One advantage of domain-specific languages is that they enable program
transformation, e.g., for optimization~\cite{mernik_heering_sloane_2005}.
The DSPy prompting framework~\cite{khattab_et_al_2023} has the
tag-line ``programming, not prompting'': it finds prompts
automatically so developers need not write them by hand.
Similarly, Vieira~\cite{li_et_al_2024} extends Prolog with LLMs as
probabilistic relations, for which it automatically derives prompts.
Both DSPy and Vieira are very high-level, but unlike PDL, both take
away control from the developer over exact prompts.
One language that lets users gradually adjust this tradeoff between
automation and control for AI pipelines is Lale~\cite{baudart_et_al_2021};
however, Lale does not focus on LLM prompting in particular.
Whereas DSPy, Vieira, and Lale optimize for predictive performance,
another objective to optimize for is computational performance.
SGLang~\cite{zheng_et_al_2023} does that by scheduling to better
leverage prefix caching (to get more cache hits in a KV
cache~\cite{kwon_et_al_2023}).
Future work will explore whether PDL's declarative nature can enable
similar computational performance optimizations.

Recently, a new crop of prompting framework has emerged that focuses
on LLM-based agents.
AutoGen~\cite{wu_et_al_2023} is a multi-agent framework where
everything consists of agents and conversations.
Other multi-agent frameworks include CrewAI~\cite{moura_2023} and
GPTSwarm~\cite{zhuge_et_al_2024}.
These frameworks prioritize support for agents over support for other
LLM-based use-cases.
While PDL supports agents as well, it aims for a more balanced stance,
where agents are just one of many prompting techniques.

\section{Conclusion}\label{sec:conclusion}
PDL is a declarative data-oriented language: a program consists of
YAML blocks, where each block either is a literal piece of data or
produces data.
The mental model is that executing a block appends its data to the
background context, and subsequent LLM calls use that context as their
prompt.
This paper introduces the language via example programs and a tour
of the grammar and tooling.
The declarative nature of the language also makes it amenable to
automatic optimizations for speed, accuracy, and security, which will
be forthcoming in future work.
PDL is ready to use and open-source at
\url{https://github.com/IBM/prompt-declaration-language}.

\bibliography{bibfile}


\begin{thebibliography}{36}

%%% the Web address.

\ifx \showCODEN    \undefined \def \showCODEN     #1{\unskip}     \fi
\ifx \showDOI      \undefined \def \showDOI       #1{#1}\fi
\ifx \showISBNx    \undefined \def \showISBNx     #1{\unskip}     \fi
\ifx \showISBNxiii \undefined \def \showISBNxiii  #1{\unskip}     \fi
\ifx \showISSN     \undefined \def \showISSN      #1{\unskip}     \fi
\ifx \showLCCN     \undefined \def \showLCCN      #1{\unskip}     \fi
\ifx \shownote     \undefined \def \shownote      #1{#1}          \fi
\ifx \showarticletitle \undefined \def \showarticletitle #1{#1}   \fi
\ifx \showURL      \undefined \def \showURL       {\relax}        \fi
% invisible to TeX
\providecommand\bibfield[2]{#2}
\providecommand\bibinfo[2]{#2}
\providecommand\natexlab[1]{#1}
\providecommand\showeprint[2][]{arXiv:#2}

\bibitem[Abdelaziz et~al\mbox{.}(2024)]%
        {abdelaziz_et_al_2024}
\bibfield{author}{\bibinfo{person}{Ibrahim Abdelaziz}, \bibinfo{person}{Kinjal
  Basu}, \bibinfo{person}{Mayank Agarwal}, \bibinfo{person}{Sadhana Kumaravel},
  \bibinfo{person}{Matthew Stallone}, \bibinfo{person}{Rameswar Panda},
  \bibinfo{person}{Yara Rizk}, \bibinfo{person}{GP Bhargav},
  \bibinfo{person}{Maxwell Crouse}, \bibinfo{person}{Chulaka Gunasekara},
  \bibinfo{person}{Shajith Ikbal}, \bibinfo{person}{Sachin Joshi},
  \bibinfo{person}{Hima Karanam}, \bibinfo{person}{Vineet Kumar},
  \bibinfo{person}{Asim Munawar}, \bibinfo{person}{Sumit Neelam},
  \bibinfo{person}{Dinesh Raghu}, \bibinfo{person}{Udit Sharma},
  \bibinfo{person}{Adriana~Meza Soria}, \bibinfo{person}{Dheeraj Sreedhar},
  \bibinfo{person}{Praveen Venkateswaran}, \bibinfo{person}{Merve Unuvar},
  \bibinfo{person}{David Cox}, \bibinfo{person}{Salim Roukos},
  \bibinfo{person}{Luis Lastras}, {and} \bibinfo{person}{Pavan Kapanipathi}.}
  \bibinfo{year}{2024}\natexlab{}.
\newblock \bibinfo{title}{Granite-Function Calling Model: Introducing Function
  Calling Abilities via Multi-task Learning of Granular Tasks}.
\newblock
\newblock
\urldef\tempurl%
\url{https://arxiv.org/abs/2407.00121}
\showURL{%
\tempurl}


\bibitem[Austin et~al\mbox{.}(2021)]%
        {austin_et_al_2021}
\bibfield{author}{\bibinfo{person}{Jacob Austin}, \bibinfo{person}{Augustus
  Odena}, \bibinfo{person}{Maxwell Nye}, \bibinfo{person}{Maarten Bosma},
  \bibinfo{person}{Henryk Michalewski}, \bibinfo{person}{David Dohan},
  \bibinfo{person}{Ellen Jiang}, \bibinfo{person}{Carrie Cai},
  \bibinfo{person}{Michael Terry}, \bibinfo{person}{Quoc Le}, {and}
  \bibinfo{person}{Charles Sutton}.} \bibinfo{year}{2021}\natexlab{}.
\newblock \bibinfo{title}{Program Synthesis with Large Language Models}.
\newblock
\newblock
\urldef\tempurl%
\url{https://arxiv.org/abs/2108.07732}
\showURL{%
\tempurl}


\bibitem[Baudart et~al\mbox{.}(2021)]%
        {baudart_et_al_2021}
\bibfield{author}{\bibinfo{person}{Guillaume Baudart}, \bibinfo{person}{Martin
  Hirzel}, \bibinfo{person}{Kiran Kate}, \bibinfo{person}{Parikshit Ram},
  \bibinfo{person}{Avraham Shinnar}, {and} \bibinfo{person}{Jason Tsay}.}
  \bibinfo{year}{2021}\natexlab{}.
\newblock \showarticletitle{Pipeline Combinators for Gradual {AutoML}}. In
  \bibinfo{booktitle}{\emph{Advances in Neural Information Processing Systems
  (NeurIPS)}}. \bibinfo{pages}{19705--19718}.
\newblock
\urldef\tempurl%
\url{https://proceedings.neurips.cc/paper/2021/file/a3b36cb25e2e0b93b5f334ffb4e4064e-Paper.pdf}
\showURL{%
\tempurl}


\bibitem[Ben-Kiki et~al\mbox{.}(2004)]%
        {benkiti_et_al_2004}
\bibfield{author}{\bibinfo{person}{Oren Ben-Kiki}, \bibinfo{person}{Clark
  Evans}, {and} \bibinfo{person}{Brian Ingerson}.}
  \bibinfo{year}{2004}\natexlab{}.
\newblock \bibinfo{title}{{YAML} Ain't Markup Language}.
\newblock
\newblock
\urldef\tempurl%
\url{http://yaml.org/spec/history/2004-12-28/2004-12-28.pdf}
\showURL{%
\tempurl}


\bibitem[Beurer-Kellner et~al\mbox{.}(2023)]%
        {beurerkellner_fischer_vechev_2023}
\bibfield{author}{\bibinfo{person}{Luca Beurer-Kellner}, \bibinfo{person}{Marc
  Fischer}, {and} \bibinfo{person}{Martin Vechev}.}
  \bibinfo{year}{2023}\natexlab{}.
\newblock \showarticletitle{Prompting Is Programming: A Query Language for
  Large Language Models}. In \bibinfo{booktitle}{\emph{Conference on
  Programming Language Design and Implementation (PLDI)}}.
  \bibinfo{pages}{1946--1969}.
\newblock
\urldef\tempurl%
\url{https://doi.org/10.1145/3591300}
\showURL{%
\tempurl}


\bibitem[Brown et~al\mbox{.}(2020)]%
        {brown_et_al_2020}
\bibfield{author}{\bibinfo{person}{Tom~B. Brown}, \bibinfo{person}{Benjamin
  Mann}, \bibinfo{person}{Nick Ryder}, \bibinfo{person}{Melanie Subbiah},
  \bibinfo{person}{Jared Kaplan}, \bibinfo{person}{Prafulla Dhariwal},
  \bibinfo{person}{Arvind Neelakantan}, \bibinfo{person}{Pranav Shyam},
  \bibinfo{person}{Girish Sastry}, \bibinfo{person}{Amanda Askell},
  \bibinfo{person}{Sandhini Agarwal}, \bibinfo{person}{Ariel Herbert-Voss},
  \bibinfo{person}{Gretchen Krueger}, \bibinfo{person}{Tom Henighan},
  \bibinfo{person}{Rewon Child}, \bibinfo{person}{Aditya Ramesh},
  \bibinfo{person}{Daniel~M. Ziegler}, \bibinfo{person}{Jeffrey Wu},
  \bibinfo{person}{Clemens Winter}, \bibinfo{person}{Christopher Hesse},
  \bibinfo{person}{Mark Chen}, \bibinfo{person}{Eric Sigler},
  \bibinfo{person}{Mateusz Litwin}, \bibinfo{person}{Scott Gray},
  \bibinfo{person}{Benjamin Chess}, \bibinfo{person}{Jack Clark},
  \bibinfo{person}{Christopher Berner}, \bibinfo{person}{Sam McCandlish},
  \bibinfo{person}{Alec Radford}, \bibinfo{person}{Ilya Sutskever}, {and}
  \bibinfo{person}{Dario Amodei}.} \bibinfo{year}{2020}\natexlab{}.
\newblock \bibinfo{title}{Language Models are Few-Shot Learners}.
\newblock
\newblock
\urldef\tempurl%
\url{https://arxiv.org/abs/2005.14165}
\showURL{%
\tempurl}


\bibitem[{Chase et al.}(2022)]%
        {chase_et_al_2022}
\bibfield{author}{\bibinfo{person}{Harrison {Chase et al.}}}
  \bibinfo{year}{2022}\natexlab{}.
\newblock \bibinfo{title}{{LangChain}}.
\newblock
\newblock
\urldef\tempurl%
\url{https://github.com/langchain-ai/langchain}
\showURL{%
\tempurl}


\bibitem[Cooper et~al\mbox{.}(2006)]%
        {cooper_et_al_2006}
\bibfield{author}{\bibinfo{person}{Ezra Cooper}, \bibinfo{person}{Sam Lindley},
  \bibinfo{person}{Philip Wadler}, {and} \bibinfo{person}{Jeremy Yallop}.}
  \bibinfo{year}{2006}\natexlab{}.
\newblock \showarticletitle{Links: Web Programming Without Tiers}. In
  \bibinfo{booktitle}{\emph{Symposium on Formal Methods for Components and
  Objects (FMCO)}}. \bibinfo{pages}{266--296}.
\newblock
\urldef\tempurl%
\url{https://doi.org/10.1007/978-3-540-74792-5_12}
\showURL{%
\tempurl}


\bibitem[Crouse et~al\mbox{.}(2024)]%
        {crouse_et_al_2024}
\bibfield{author}{\bibinfo{person}{Maxwell Crouse}, \bibinfo{person}{Ibrahim
  Abdelaziz}, \bibinfo{person}{Ramon Astudillo}, \bibinfo{person}{Kinjal Basu},
  \bibinfo{person}{Soham Dan}, \bibinfo{person}{Sadhana Kumaravel},
  \bibinfo{person}{Achille Fokoue}, \bibinfo{person}{Pavan Kapanipathi},
  \bibinfo{person}{Salim Roukos}, {and} \bibinfo{person}{Luis Lastras}.}
  \bibinfo{year}{2024}\natexlab{}.
\newblock \bibinfo{title}{Formally Specifying the High-Level Behavior of
  {LLM}-Based Agents}.
\newblock
\newblock
\urldef\tempurl%
\url{https://arxiv.org/abs/2310.08535}
\showURL{%
\tempurl}


\bibitem[Gao et~al\mbox{.}(2023)]%
        {gao_et_al_2023}
\bibfield{author}{\bibinfo{person}{Luyu Gao}, \bibinfo{person}{Aman Madaan},
  \bibinfo{person}{Shuyan Zhou}, \bibinfo{person}{Uri Alon},
  \bibinfo{person}{Pengfei Liu}, \bibinfo{person}{Yiming Yang},
  \bibinfo{person}{Jamie Callan}, {and} \bibinfo{person}{Graham Neubig}.}
  \bibinfo{year}{2023}\natexlab{}.
\newblock \showarticletitle{{PAL}: Program-aided Language Models}. In
  \bibinfo{booktitle}{\emph{International Conference on Machine Learning
  (ICML)}}. \bibinfo{pages}{10764--10799}.
\newblock
\urldef\tempurl%
\url{https://proceedings.mlr.press/v202/gao23f.html}
\showURL{%
\tempurl}


\bibitem[{Granite Team, IBM}(2024)]%
        {granite_3_0}
\bibfield{author}{\bibinfo{person}{{Granite Team, IBM}}.}
  \bibinfo{year}{2024}\natexlab{}.
\newblock \bibinfo{title}{Granite 3.0 Language Models}.
\newblock
\newblock
\urldef\tempurl%
\url{https://github.com/ibm-granite/granite-3.0-language-models/blob/main/paper.pdf}
\showURL{%
\tempurl}


\bibitem[{Hugging Face}(2023)]%
        {huggingface_2023}
\bibfield{author}{\bibinfo{person}{{Hugging Face}}.}
  \bibinfo{year}{2023}\natexlab{}.
\newblock \bibinfo{title}{Chat Templates}.
\newblock
\newblock
\urldef\tempurl%
\url{https://huggingface.co/docs/transformers/en/chat_templating}
\showURL{%
\tempurl}


\bibitem[{IBM}(2023)]%
        {ibm_2023}
\bibfield{author}{\bibinfo{person}{{IBM}}.} \bibinfo{year}{2023}\natexlab{}.
\newblock \bibinfo{title}{watsonx}.
\newblock
\newblock
\urldef\tempurl%
\url{https://www.ibm.com/watsonx}
\showURL{%
\tempurl}


\bibitem[Jimenez et~al\mbox{.}(2024)]%
        {jimenez_et_al_2024}
\bibfield{author}{\bibinfo{person}{Carlos~E. Jimenez}, \bibinfo{person}{John
  Yang}, \bibinfo{person}{Alexander Wettig}, \bibinfo{person}{Shunyu Yao},
  \bibinfo{person}{Kexin Pei}, \bibinfo{person}{Ofir Press}, {and}
  \bibinfo{person}{Karthik Narasimhan}.} \bibinfo{year}{2024}\natexlab{}.
\newblock \showarticletitle{{SWE}-bench: Can Language Models Resolve Real-World
  {GitHub} Issues?}. In \bibinfo{booktitle}{\emph{International Conference on
  Learning Representations (ICLR)}}.
\newblock
\urldef\tempurl%
\url{https://openreview.net/forum?id=VTF8yNQM66}
\showURL{%
\tempurl}


\bibitem[Khattab et~al\mbox{.}(2023)]%
        {khattab_et_al_2023}
\bibfield{author}{\bibinfo{person}{Omar Khattab}, \bibinfo{person}{Arnav
  Singhvi}, \bibinfo{person}{Paridhi Maheshwari}, \bibinfo{person}{Zhiyuan
  Zhang}, \bibinfo{person}{Keshav Santhanam}, \bibinfo{person}{Sri
  Vardhamanan}, \bibinfo{person}{Saiful Haq}, \bibinfo{person}{Ashutosh
  Sharma}, \bibinfo{person}{Thomas~T. Joshi}, \bibinfo{person}{Hanna Moazam},
  \bibinfo{person}{Heather Miller}, \bibinfo{person}{Matei Zaharia}, {and}
  \bibinfo{person}{Christopher Potts}.} \bibinfo{year}{2023}\natexlab{}.
\newblock \bibinfo{title}{{DSPy}: Compiling Declarative Language Model Calls
  into Self-Improving Pipelines}.
\newblock
\newblock
\urldef\tempurl%
\url{https://arxiv.org/abs/2310.03714}
\showURL{%
\tempurl}


\bibitem[Krishna et~al\mbox{.}(2024)]%
        {krishna_et_al_2024}
\bibfield{author}{\bibinfo{person}{Rahul Krishna}, \bibinfo{person}{Rangeet
  Pan}, \bibinfo{person}{Raju Pavuluri}, \bibinfo{person}{Srikanth
  Tamilselvam}, \bibinfo{person}{Maja Vukovic}, {and} \bibinfo{person}{Saurabh
  Sinha}.} \bibinfo{year}{2024}\natexlab{}.
\newblock \bibinfo{title}{{Codellm-Devkit}: A Framework for Contextualizing
  Code {LLMs} with Program Analysis Insights}.
\newblock
\newblock
\urldef\tempurl%
\url{https://arxiv.org/abs/2410.13007}
\showURL{%
\tempurl}


\bibitem[Kwon et~al\mbox{.}(2023)]%
        {kwon_et_al_2023}
\bibfield{author}{\bibinfo{person}{Woosuk Kwon}, \bibinfo{person}{Zhuohan Li},
  \bibinfo{person}{Siyuan Zhuang}, \bibinfo{person}{Ying Sheng},
  \bibinfo{person}{Lianmin Zheng}, \bibinfo{person}{Cody~Hao Yu},
  \bibinfo{person}{Joseph Gonzalez}, \bibinfo{person}{Hao Zhang}, {and}
  \bibinfo{person}{Ion Stoica}.} \bibinfo{year}{2023}\natexlab{}.
\newblock \showarticletitle{Efficient Memory Management for Large Language
  Model Serving with {PagedAttention}}. In \bibinfo{booktitle}{\emph{Symposium
  on Operating Systems Principles (SOSP)}}. \bibinfo{pages}{611--626}.
\newblock
\urldef\tempurl%
\url{https://doi.org/10.1145/3600006.3613165}
\showURL{%
\tempurl}


\bibitem[Lewis et~al\mbox{.}(2020)]%
        {lewis_et_al_2020}
\bibfield{author}{\bibinfo{person}{Patrick Lewis}, \bibinfo{person}{Ethan
  Perez}, \bibinfo{person}{Aleksandra Piktus}, \bibinfo{person}{Fabio Petroni},
  \bibinfo{person}{Vladimir Karpukhin}, \bibinfo{person}{Naman Goyal},
  \bibinfo{person}{Heinrich K{\"u}ttler}, \bibinfo{person}{Mike Lewis},
  \bibinfo{person}{Wen-tau Yih}, \bibinfo{person}{Tim Rockt{\"a}schel},
  \bibinfo{person}{Sebastian Riedel}, {and} \bibinfo{person}{Douwe Kiela}.}
  \bibinfo{year}{2020}\natexlab{}.
\newblock \showarticletitle{Retrieval-Augmented Generation for
  Knowledge-Intensive {NLP} Tasks}. In \bibinfo{booktitle}{\emph{Conference on
  Neural Information Processing Systems (NeurIPS)}}.
  \bibinfo{pages}{9459--9474}.
\newblock
\urldef\tempurl%
\url{https://proceedings.neurips.cc/paper/2020/hash/6b493230205f780e1bc26945df7481e5-Abstract.html}
\showURL{%
\tempurl}


\bibitem[Li et~al\mbox{.}(2024)]%
        {li_et_al_2024}
\bibfield{author}{\bibinfo{person}{Ziyang Li}, \bibinfo{person}{Jiani Huang},
  \bibinfo{person}{Jason Liu}, \bibinfo{person}{Felix Zhu},
  \bibinfo{person}{Eric Zhao}, \bibinfo{person}{William Dodds},
  \bibinfo{person}{Neelay Velingker}, \bibinfo{person}{Rajeev Alur}, {and}
  \bibinfo{person}{Mayur Naik}.} \bibinfo{year}{2024}\natexlab{}.
\newblock \showarticletitle{Relational Programming with Foundation Models}. In
  \bibinfo{booktitle}{\emph{Conference on Artificial Intelligence (AAAI)}}.
  \bibinfo{pages}{10635--10644}.
\newblock
\urldef\tempurl%
\url{https://doi.org/10.1609/aaai.v38i9.28934}
\showURL{%
\tempurl}


\bibitem[Liu et~al\mbox{.}(2023)]%
        {liu_et_al_2023}
\bibfield{author}{\bibinfo{person}{Xiaoxia Liu}, \bibinfo{person}{Jingyi Wang},
  \bibinfo{person}{Jun Sun}, \bibinfo{person}{Xiaohan Yuan},
  \bibinfo{person}{Guoliang Dong}, \bibinfo{person}{Peng Di},
  \bibinfo{person}{Wenhai Wang}, {and} \bibinfo{person}{Dongxia Wang}.}
  \bibinfo{year}{2023}\natexlab{}.
\newblock \bibinfo{title}{Prompting Frameworks for Large Language Models: A
  Survey}.
\newblock
\newblock
\urldef\tempurl%
\url{https://arxiv.org/abs/2311.12785}
\showURL{%
\tempurl}


\bibitem[McCarthy(1960)]%
        {mccarthy_1960}
\bibfield{author}{\bibinfo{person}{John McCarthy}.}
  \bibinfo{year}{1960}\natexlab{}.
\newblock \showarticletitle{Recursive functions of symbolic expressions and
  their computation by machine, Part I}.
\newblock \bibinfo{journal}{\emph{Communications of the ACM (CACM)}}
  \bibinfo{volume}{3}, \bibinfo{number}{4} (\bibinfo{date}{April}
  \bibinfo{year}{1960}), \bibinfo{pages}{184--195}.
\newblock
\urldef\tempurl%
\url{https://doi.org/10.1145/367177.367199}
\showURL{%
\tempurl}


\bibitem[Mernik et~al\mbox{.}(2005)]%
        {mernik_heering_sloane_2005}
\bibfield{author}{\bibinfo{person}{Marjan Mernik}, \bibinfo{person}{Jan
  Heering}, {and} \bibinfo{person}{Anthony~M. Sloane}.}
  \bibinfo{year}{2005}\natexlab{}.
\newblock \showarticletitle{When and how to develop domain-specific languages}.
\newblock \bibinfo{journal}{\emph{ACM Computing Surveys (CSUR)}}
  \bibinfo{volume}{37}, \bibinfo{number}{4} (\bibinfo{year}{2005}),
  \bibinfo{pages}{316--344}.
\newblock
\urldef\tempurl%
\url{https://doi.org/10.1145/1118890.1118892}
\showURL{%
\tempurl}


\bibitem[Microsoft(2023)]%
        {microsoft_2023}
\bibfield{author}{\bibinfo{person}{Microsoft}.}
  \bibinfo{year}{2023}\natexlab{}.
\newblock \bibinfo{title}{\{guidance\}: A guidance language for controlling
  large language models}.
\newblock
\newblock
\urldef\tempurl%
\url{https://github.com/langchain-ai/langchain}
\showURL{%
\tempurl}


\bibitem[Moura(2023)]%
        {moura_2023}
\bibfield{author}{\bibinfo{person}{Jo{\~a}o Moura}.}
  \bibinfo{year}{2023}\natexlab{}.
\newblock \bibinfo{title}{{CrewAI}: Framework for orchestrating role-playing,
  autonomous {AI} agents}.
\newblock
\newblock
\urldef\tempurl%
\url{https://github.com/crewAIInc/crewAI}
\showURL{%
\tempurl}


\bibitem[Pezoa et~al\mbox{.}(2016)]%
        {pezoa_et_al_2016}
\bibfield{author}{\bibinfo{person}{Felipe Pezoa}, \bibinfo{person}{Juan~L.
  Reutter}, \bibinfo{person}{Fernando Suarez}, \bibinfo{person}{Mart\'{\i}n
  Ugarte}, {and} \bibinfo{person}{Domagoj Vrgo\v{c}}.}
  \bibinfo{year}{2016}\natexlab{}.
\newblock \showarticletitle{Foundations of {JSON} Schema}. In
  \bibinfo{booktitle}{\emph{International Conference on World Wide Web (WWW)}}.
  \bibinfo{pages}{263--273}.
\newblock
\urldef\tempurl%
\url{https://doi.org/10.1145/2872427.2883029}
\showURL{%
\tempurl}


\bibitem[Ronacher(2008)]%
        {ronacher_2008}
\bibfield{author}{\bibinfo{person}{Armin Ronacher}.}
  \bibinfo{year}{2008}\natexlab{}.
\newblock \bibinfo{title}{Jinja2 Documentation Release 2.0}.
\newblock
\newblock
\urldef\tempurl%
\url{http://mitsuhiko.pocoo.org/jinja2docs/Jinja2.pdf}
\showURL{%
\tempurl}


\bibitem[Rush(2023)]%
        {rush_2023}
\bibfield{author}{\bibinfo{person}{Alexander Rush}.}
  \bibinfo{year}{2023}\natexlab{}.
\newblock \showarticletitle{{MiniChain}: A Small Library for Coding with Large
  Language Models}. In \bibinfo{booktitle}{\emph{Conference on Empirical
  Methods in Natural Language Processing: System Demonstrations (EMNLP-Demo)}}.
  \bibinfo{pages}{311--317}.
\newblock
\urldef\tempurl%
\url{https://aclanthology.org/2023.emnlp-demo.27/}
\showURL{%
\tempurl}


\bibitem[Schick et~al\mbox{.}(2023)]%
        {schick_et_al_2023}
\bibfield{author}{\bibinfo{person}{Timo Schick}, \bibinfo{person}{Jane
  Dwivedi-Yu}, \bibinfo{person}{Roberto Dessi}, \bibinfo{person}{Roberta
  Raileanu}, \bibinfo{person}{Maria Lomeli}, \bibinfo{person}{Luke
  Zettlemoyer}, \bibinfo{person}{Nicola Cancedda}, {and}
  \bibinfo{person}{Thomas Scialom}.} \bibinfo{year}{2023}\natexlab{}.
\newblock \showarticletitle{Toolformer: Language Models Can Teach Themselves to
  Use Tools}. In \bibinfo{booktitle}{\emph{Advances in Neural Information
  Processing Systems (NeurIPS)}}.
\newblock
\urldef\tempurl%
\url{https://proceedings.neurips.cc/paper_files/paper/2023/hash/d842425e4bf79ba039352da0f658a906-Abstract-Conference.html}
\showURL{%
\tempurl}


\bibitem[Scholak et~al\mbox{.}(2021)]%
        {scholak_schucher_bahdanau_2021}
\bibfield{author}{\bibinfo{person}{Torsten Scholak}, \bibinfo{person}{Nathan
  Schucher}, {and} \bibinfo{person}{Dzmitry Bahdanau}.}
  \bibinfo{year}{2021}\natexlab{}.
\newblock \showarticletitle{{PICARD}: Parsing Incrementally for Constrained
  Auto-Regressive Decoding from Language Models}. In
  \bibinfo{booktitle}{\emph{Conference on Empirical Methods in Natural Language
  Processing (EMNLP)}}. \bibinfo{pages}{9895--9901}.
\newblock
\urldef\tempurl%
\url{https://doi.org/10.18653/v1/2021.emnlp-main.779}
\showURL{%
\tempurl}


\bibitem[{van Wijngaarden} et~al\mbox{.}(1977)]%
        {vanwijngaarden_et_al_1977}
\bibfield{author}{\bibinfo{person}{A. {van Wijngaarden}}, \bibinfo{person}{B.J.
  Mailloux}, \bibinfo{person}{J.E.L. Peck}, \bibinfo{person}{C.H.A. Koster},
  \bibinfo{person}{M. Sintzoff}, \bibinfo{person}{C.H. Lindsey},
  \bibinfo{person}{L.G.T. Meertens}, {and} \bibinfo{person}{R.G. Fisker}.}
  \bibinfo{year}{1977}\natexlab{}.
\newblock \showarticletitle{Revised Report on the Algorithmic Language {ALGOL}
  68}.
\newblock \bibinfo{journal}{\emph{ACM SIGPLAN Notices}} \bibinfo{volume}{12},
  \bibinfo{number}{5} (\bibinfo{date}{May} \bibinfo{year}{1977}),
  \bibinfo{pages}{1--70}.
\newblock
\urldef\tempurl%
\url{https://doi.org/10.1145/954652.1781176}
\showURL{%
\tempurl}


\bibitem[White et~al\mbox{.}(2023)]%
        {white_et_al_2023}
\bibfield{author}{\bibinfo{person}{Jules White}, \bibinfo{person}{Quchen Fu},
  \bibinfo{person}{Sam Hays}, \bibinfo{person}{Michael Sandborn},
  \bibinfo{person}{Carlos Olea}, \bibinfo{person}{Henry Gilbert},
  \bibinfo{person}{Ashraf Elnashar}, \bibinfo{person}{Jesse Spencer-Smith},
  {and} \bibinfo{person}{Douglas~C. Schmidt}.} \bibinfo{year}{2023}\natexlab{}.
\newblock \bibinfo{title}{A Prompt Pattern Catalog to Enhance Prompt
  Engineering with {ChatGPT}}.
\newblock
\newblock
\urldef\tempurl%
\url{https://arxiv.org/abs/2302.11382}
\showURL{%
\tempurl}


\bibitem[Wu et~al\mbox{.}(2023)]%
        {wu_et_al_2023}
\bibfield{author}{\bibinfo{person}{Qingyun Wu}, \bibinfo{person}{Gagan Bansal},
  \bibinfo{person}{Jieyu Zhang}, \bibinfo{person}{Yiran Wu},
  \bibinfo{person}{Beibin Li}, \bibinfo{person}{Erkang Zhu},
  \bibinfo{person}{Li Jiang}, \bibinfo{person}{Xiaoyun Zhang},
  \bibinfo{person}{Shaokun Zhang}, \bibinfo{person}{Jiale Liu},
  \bibinfo{person}{Ahmed~Hassan Awadallah}, \bibinfo{person}{Ryen~W White},
  \bibinfo{person}{Doug Burger}, {and} \bibinfo{person}{Chi Wang}.}
  \bibinfo{year}{2023}\natexlab{}.
\newblock \bibinfo{title}{{AutoGen}: Enabling Next-Gen {LLM} Applications via
  Multi-Agent Conversation}.
\newblock
\newblock
\urldef\tempurl%
\url{https://arxiv.org/abs/2308.08155}
\showURL{%
\tempurl}


\bibitem[Xu et~al\mbox{.}(2023)]%
        {xu_et_al_2023}
\bibfield{author}{\bibinfo{person}{Binfeng Xu}, \bibinfo{person}{Zhiyuan Peng},
  \bibinfo{person}{Bowen Lei}, \bibinfo{person}{Subhabrata Mukherjee}, {and}
  \bibinfo{person}{Dongkuan Xu}.} \bibinfo{year}{2023}\natexlab{}.
\newblock \bibinfo{title}{Decoupling Reasoning from Observations for Efficient
  Augmented Language Models}.
\newblock
\newblock
\urldef\tempurl%
\url{https://openreview.net/forum?id=CpgoO6j6W1}
\showURL{%
\tempurl}


\bibitem[Yao et~al\mbox{.}(2023)]%
        {yao_et_al_2023}
\bibfield{author}{\bibinfo{person}{Shunyu Yao}, \bibinfo{person}{Jeffrey Zhao},
  \bibinfo{person}{Dian Yu}, \bibinfo{person}{Nan Du}, \bibinfo{person}{Izhak
  Shafran}, \bibinfo{person}{Karthik~R Narasimhan}, {and} \bibinfo{person}{Yuan
  Cao}.} \bibinfo{year}{2023}\natexlab{}.
\newblock \showarticletitle{{ReAct}: Synergizing Reasoning and Acting in
  Language Models}. In \bibinfo{booktitle}{\emph{International Conference on
  Learning Representations (ICLR)}}.
\newblock
\urldef\tempurl%
\url{https://openreview.net/forum?id=WE_vluYUL-X}
\showURL{%
\tempurl}


\bibitem[Zheng et~al\mbox{.}(2023)]%
        {zheng_et_al_2023}
\bibfield{author}{\bibinfo{person}{Lianmin Zheng}, \bibinfo{person}{Liangsheng
  Yin}, \bibinfo{person}{Zhiqiang Xie}, \bibinfo{person}{Jeff Huang},
  \bibinfo{person}{Chuyue Sun}, \bibinfo{person}{Cody~Hao Yu},
  \bibinfo{person}{Shiyi Cao}, \bibinfo{person}{Christos Kozyrakis},
  \bibinfo{person}{Ion Stoica}, \bibinfo{person}{Joseph~E. Gonzalez},
  \bibinfo{person}{Clark Barrett}, {and} \bibinfo{person}{Ying Sheng}.}
  \bibinfo{year}{2023}\natexlab{}.
\newblock \bibinfo{title}{Efficiently Programming Large Language Models using
  {SGLang}}.
\newblock
\newblock
\urldef\tempurl%
\url{https://arxiv.org/abs/2312.07104}
\showURL{%
\tempurl}


\bibitem[Zhuge et~al\mbox{.}(2024)]%
        {zhuge_et_al_2024}
\bibfield{author}{\bibinfo{person}{Mingchen Zhuge}, \bibinfo{person}{Wenyi
  Wang}, \bibinfo{person}{Louis Kirsch}, \bibinfo{person}{Francesco Faccio},
  \bibinfo{person}{Dmitrii Khizbullin}, {and} \bibinfo{person}{J{\"u}rgen
  Schmidhuber}.} \bibinfo{year}{2024}\natexlab{}.
\newblock \showarticletitle{{GPTSwarm}: Language Agents as Optimizable Graphs}.
  In \bibinfo{booktitle}{\emph{International Conference on Machine Learning
  (ICML)}}.
\newblock
\urldef\tempurl%
\url{https://openreview.net/forum?id=uTC9AFXIhg}
\showURL{%
\tempurl}


\end{thebibliography}

\end{document}